\def\year{2021}\relax
\documentclass[letterpaper]{article} 
\usepackage{aaai21}  
\usepackage{times}  
\usepackage{helvet} 
\usepackage{courier}  
\usepackage[hyphens]{url}  
\usepackage{graphicx} 
\usepackage{natbib}  
\usepackage{caption} 
\usepackage{booktabs}
\usepackage{amsmath}
\usepackage{amssymb}
\usepackage{overpic}
\usepackage[switch]{lineno}
\title{ HR-Depth : High Resolution Self-Supervised Monocular Depth Estimation }
\author {
    Xiaoyang Lyu \textsuperscript{\rm 1},
    Liang Liu \textsuperscript{\rm 1},
    Mengmeng Wang \textsuperscript{\rm 1},
    Xin Kong \textsuperscript{\rm 1}, \\
    Lina Liu \textsuperscript{\rm 1},
    Yong Liu\footnote{Corresponding author} \textsuperscript{\rm 1},
    Xinxin Chen \textsuperscript{\rm 1},
    Yi Yuan \textsuperscript{\rm 2} \\
}
\affiliations{
    \textsuperscript{\rm 1} Zhejiang University \textsuperscript{\rm 2} Fuxi AI Lab, NetEase \\
    \{shawlyu, leonliuz, mengmengwang, xinkong, linaliu\}@zju.edu.cn, \\
    yongliu@iipc.zju.edu.cn, cxx-compiler@zju.edu.cn, yuanyi@corp.netease.com
}

\begin{document}
\maketitle

\begin{abstract}
    Self-supervised learning shows great potential in monocular depth estimation, using image sequences as the only source of supervision. Although people
    try to use the high-resolution image for depth estimation, the accuracy of prediction has not been significantly improved. In this work, we find the core reason
    comes from the inaccurate depth estimation in large gradient regions, making the bilinear interpolation error gradually disappear as the resolution increases.
    To obtain more accurate depth estimation in large gradient regions, it is necessary to obtain high-resolution features with spatial and semantic information. 
    Therefore, we present an improved DepthNet, HR-Depth, with two effective strategies: (1) redesign the skip-connection in DepthNet to get better high-resolution
    features and (2) propose feature fusion Squeeze-and-Excitation(fSE) module to fuse feature more efficiently. Using Resnet-18 as the encoder, HR-Depth surpasses 
    all previous state-of-the-art(SoTA) methods with the least parameters at both high and low resolution. Moreover, previous state-of-the-art methods are based on fairly
    complex and deep networks with a mass of parameters which limits their real applications. Thus we also construct a lightweight network which uses MobileNetV3 as
    encoder. Experiments show that the lightweight network can perform on par with many large models like Monodepth2 at high-resolution with only $20\%$ parameters.
    All codes and models will be available at \url{https://github.com/shawLyu/HR-Depth}.
\end{abstract}
\section{Introduction}
Accurate depth estimation from single image is an active research filed to help computer reconstruct and understand the real scenes. It also has a large range of 
applications in diverse fields such as autonomous vehicles, robotics, augmented reality, etc. While supervised monocular depth estimation has been successful, it is 
suffering from the expensive access to ground truth. Self-supervised methods use geometrical constraints on image sequences or stereo images as the sole source of 
supervision. 

Recent works \citep{zhou2017unsupervised, godard2017unsupervised} in self-supervised depth estimation are limited to training in low-resolution inputs due to the large memory
requirements of the model. However, with the improvement of computing and storage capacity, high-resolution images are used by more and more computer vision tasks. In depth
estimation, \cite{superdepth} introduce sub-pixel-convolutional \cite{shi2016real-time} layers replacing the deconvolution and resize-convolution layers to improve the effect of up-sampling. And 
\cite{godard2019digging} directly leverage high-resolution images for depth estimation. Although the above works explore high resolution depth estimation, but the performance 
has not improved significantly on KITTI. Inspired by this observation, we perform an analysis of existing method and find the core reason comes from the inaccurate depth 
estimation at object boundaries. 

\begin{figure}[t]
    \centering
    \includegraphics[width=1\columnwidth]{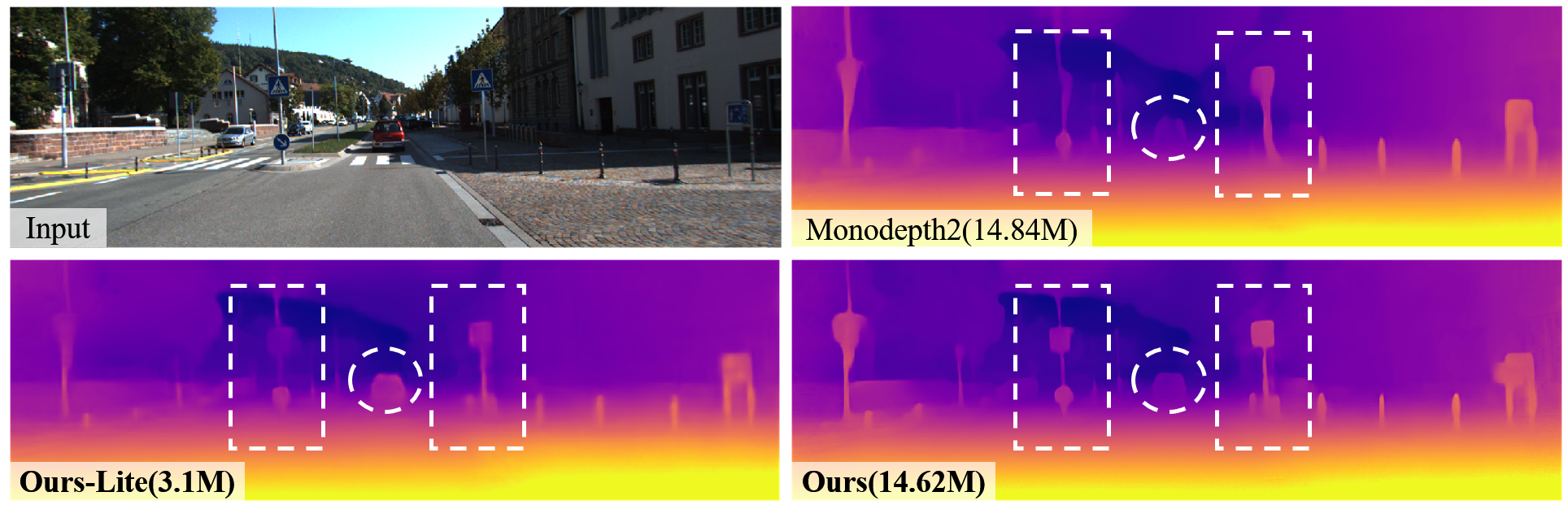} 
    \caption{\textbf{Depth prediction from single image on KITTI.} We compared our method with Monodepth2\cite{godard2019digging}, our method can 
    predict higher quality and sharper depth map with fewer parameters. The input resolution for all three models is $1024\times320$.}
\label{fig1}
\end{figure}

In depth estimation, object boundaries are mainly determined by two parts: semantic and spatial information. Semantic information obtains clear boundaries by constraining the 
categories of pixels, while spatial information uses geometric constraints to describe the outline of objects. In the previous work, the depth estimation network is based on
the U-Net\cite{ronneberger2015u} architecture, which mainly uses skip-connection to fuse semantic information and spatial information. However, the semantic gap between the encoder and decoder feature
maps is too large, which leads to a poor integration of spatial and semantic information. So the previous work is difficult to get an accurate depth estimation at object boundaries.
In order to reduce the semantic gap, we redesigned the skip connection to better fuse feature maps. Besides, we also found that the basic convolution can not integrate
spatial information and semantic information well, so we propose to replace it by a feature fusion squeeze and excitation (fSE) block. This block not only improves the feature 
fusion effect but also reduces the number of parameters. We evaluate our results on KITTI benchmark, and the experiments demonstrate that the our well-designed network can predict 
sharper edges and achieve (SoTA). 

As a side effect, higher resolution input brings extra computational costs. As a consequence, lightweight is one of the key principals for high-resolution model design. However, the previous 
SoTA has a huge amount of parameters like Packnet-SfM with $127M$ parameters, and the performance of lightweight networks is not appropriate for actual application 
like \cite{poggi2018towards}. Therefore, in order to maintain high performance of lightweight network, we introduce a simple yet effective designing strategy in this paper. We successfully 
train a lightweight network based on this strategy, which can achieve the accuracy of Monodepth2 with only $3.1M$ parameters.

To summarize, the main contributions of this work are listed below in fourfold:
\begin{itemize}
    \item We provide a deep analysis of high-resolution monocular depth estimation and prove that predicting more accurate boundaries can improve performance.
    \item We redesign the skip connection to get high-resolution semantic feature maps, which can help network predict sharper edges.
    \item We propose feature fusion squeeze and excitation block to improve the efficiency and effect of feature fusion.
    \item We present a simple yet effective lightweight design strategy to train a lightweight depth estimation network that can achieve the performance of complex network.
\end{itemize}

\begin{figure*}[ht]
    \centering
    \includegraphics[width=0.95\textwidth]{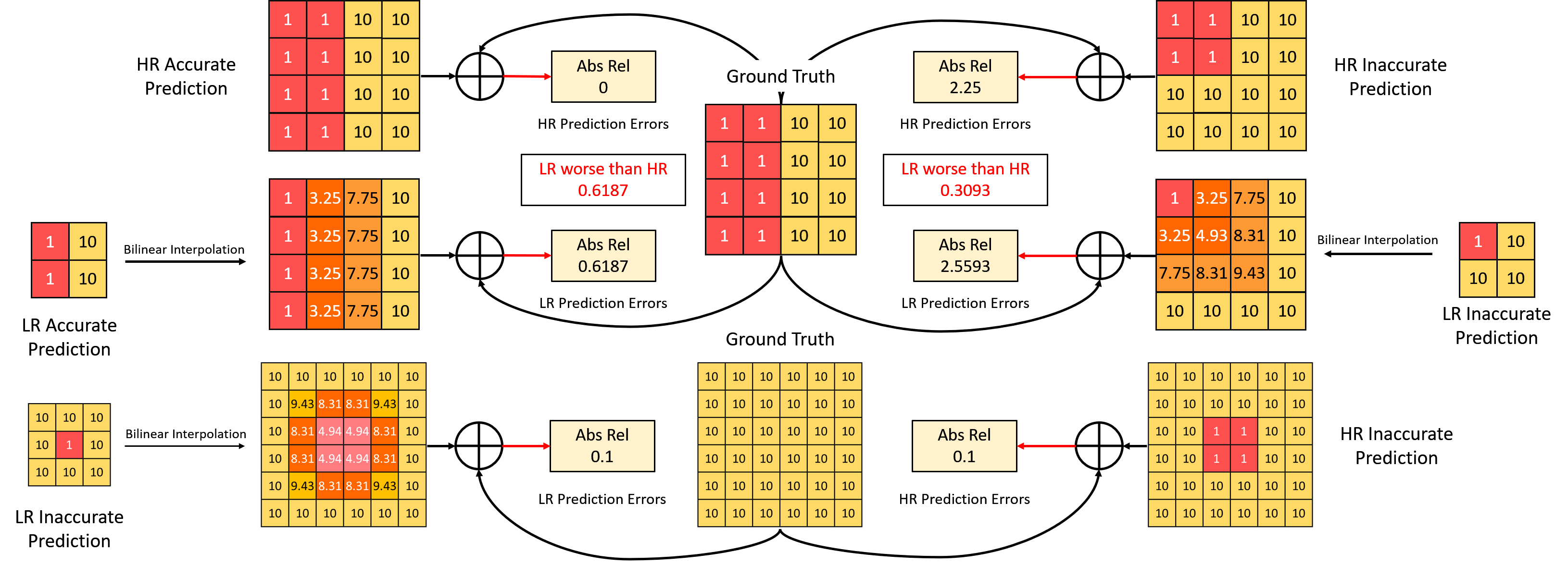} 
    \caption{\textbf{Analysis of High Resolution Depth Estimation.}  \textbf{Abs Rel} is an evaluation index in depth estimation, and lower is better. \textbf{HR} means high resolution and \textbf{LR} 
    means low resolution. All interpolation results are caculated by OpenCV library.}
\label{fig2}
\end{figure*}

\section{Related Work}
\subsection{Supervised Monocular Depth Estimation}
Depth estimation from a single image is an ill-posed problem as the same input image can be projected to multiple plausible depths. Therefore, many works begin with supervised learning. 
\cite{eigen2015predicting} is the first to propose learning-based depth estimation structure trained on RGB-D dataset. Their work considers depth estimation as a regression problem, using a coarse-to-fine network to derive pixel-by-pixel depth value.
With the rise of fully convolution neural network, \cite{laina2016deeper} uses convolution layer to replace fully connected layer, and uses pre-trained encoder for feature extraction. This work enables the depth estimation task to be trained with a deeper network
and has the accuracy comparable to the depth sensor. In order to predict sharp and accurate occlusion boundaries, \cite{ramamonjisoa2020predicting} introduces refine net predicting an additive residual depth map to refine the first estimation results. 
However, supervised methods fall into bottleneck on account of the poor generalization performance and the difficulty of obtaining ground truth depth value. So people began to explore self-supervised monocular depth estimation. 
\subsection{Self-Supervised Monocular Depth Estimation}
Using stereo images to train depth network is one intuitive of self-supervision. \cite{garg2016unsupervised} proposes one of the earliest work in self-supervised depth estimation using stereo pairs and \citep{godard2017unsupervised} produces
results superior to contemporary supervised method by introducing a left-right depth consistency loss. In order to reduce the limitation of stereo camera, \cite{zhou2017unsupervised} firstly proposes to use PoseCNN to estimate relative pose between
adjacent frames. This work enables the network to be trained completely depending on monocular image sequences. \cite{godard2019digging} introduces auto-masking and min re-projection loss to solve the problems of moving objects and occlusion and makes Monodepth2 become the most widely
used baseline. For the sake of further improving the network performance, \cite{packnet-semguided} introduces the pre-trained semantic segmentation network and pixel-adaptive convolution to guide depth network to further utilize semantic information. But there are two disadvantages 
to \cite{packnet-semguided}. First of all, we hope to relieve the pressure of pixel level annotation by self-supervising, but we need expensive semantic label in this work. Secondly, semantic segmentation and depth
estimation network should run simultaneously. It will increase the cost of depth estimation. In addition, \cite{packnet} utilizes packing and unpacking block to preserve the spatial information in image and low level feature.
They thought standard convolutional and pooling operation can not preserve sufficient details, so they proposed 3D packing and unpacking blocks to replace standard down-sample and up-sample operation. 
Packing and unpacking block are invertible, so they can recover important spatial information for depth estimation. But these two blocks depend on 3D convolution, so the number of network parameters is greatly increased and it is difficult to deploy to
mobile devices. But these two works show that abundant semantic and spatial information can get sharper edges, so as to improve the accuracy of depth estimation.

In our work, we show that, by simply fusing the information extracted by the encoder, we can obtain the ideal features with spatial and semantic information. Through our insightful design, without introducing much more parameters, these features significantly improve the overall performance, obtaining sharper edges.
\subsection{Lightweight Network for Depth Estimation}
Depth estimation from a single image is also a very attractive technique with several implications in robotic, autonomous navigation. Therefore, it is necessary to leverage depth prediction network to quickly infer an accurate depth map on a CPU.
\cite{wofk2019fastdepth} proposed a lightweight supervised depth estimation network, who use encoder-decoder architecture and include $1.34M$ parameters after pruning. In unsupervised filed, \cite{poggi2018towards,aleotti2020real} proposed PyDNet with $1.9M$ parameters. 
Although the above two works have less parameters, their performance also decreases a lot. In this paper, we propose a novel yet simple strategy for network designing, which can make the performance of lightweight network comparable to or even surpass the complex network.

\section{HR-Depth Network}
\subsection{Problem Formulation}
In self-supervised monocular depth estimation task, the goal is to use depth network $f_D$ to learn depth information $D$ from RGB image $I$. Due to the lack of ground truth depth value, we need
an additional network $f_P$ to predict relative pose $p=[R|t]$ between source image $I_s$ and target image $I_t$. As a common setting, the target image is $I_t$, and the source images set are consist of adjacent images $I_{t-1}, I_{t+1}$.
The depth network is optimized by minimizing the photometric re-projection error:
\begin{equation}
    r(I_t, I_{t'}) = \frac{\alpha}{2}(1 - SSIM(I_t, I_{t'})) + (1 - \alpha)||I_t - I_{t'}||_1
\end{equation}
where $I_{t'}$ is the warped result from $I_s$ to $I_t$, $SSIM$ is the operator of structural similarity to measure the patch similarity. We follow the form of per pixel minimum loss in \cite{godard2019digging} to handle occlusion.
The photometric loss is denoted as
\begin{equation}
    L_{re} = \min\limits_{t'}r(I_t, I_{t'}).
\end{equation}
Furthermore, in order to regularize the disparities in texture-less low-image gradient regions, edges aware smooth regularization term is used: 
\begin{equation}
    L_{smooth} = |\delta_xD_t|e^{-|\delta_xI_t|} + |\delta_yD_t|e^{-|\delta_yI_t|}.
\end{equation} 
where $\delta_x$ and $\delta_y$ symbols for partial differentiation of depth and RGB images, and the exponential operation of a matrix is an element-wise operation.
So the final loss function is the summation of the re-projected losses and the smooth loss on multi scale images:
\begin{equation}
    L_{final} = \frac{1}{s}\sum_i^s(L^{i}_{re} + \lambda L^{i}_{smooth})
\end{equation}
where $s$ is the number of scales, and $\lambda$ is the weight for smooth term.

\subsection{Analysis on High Resolution Performance}
As a consensus of dense prediction tasks, higher resolution brings more accurate results instinctively \cite{Sun_2019_CVPR, zhou2019unsupervised}. Especially in the depth estimation task, pixel level disparity is more important since 
it is inversely proportional to the square of depth error\cite{you2019pseudo}. However, we noticed that most of the previous work use low-resolution inputs and interpolate low 
resolution prediction to high-resolution one. The low-resolution experimental setup essentially makes these works cannot benefit from high-resolution images. Some methods also 
conduct high-resolution experiments where they train their models with larger images \cite{godard2019digging, superdepth}. However, the performance improvements upon the small 
inputs are very limited, i.e. \cite{superdepth}, their models cannot take advantage of high resolution. For instance, we evaluate a recent well-known work named 
Monodepth2\cite{godard2019digging} with higher resolution setting. As shown in Table 1, the depth errors are almost the same with high and low-resolution settings. 
Therefore, we argue that their method can not make full use of the information from higher input resolution.

\begin{figure*}[ht]
    \centering
    \includegraphics[width=1\textwidth]{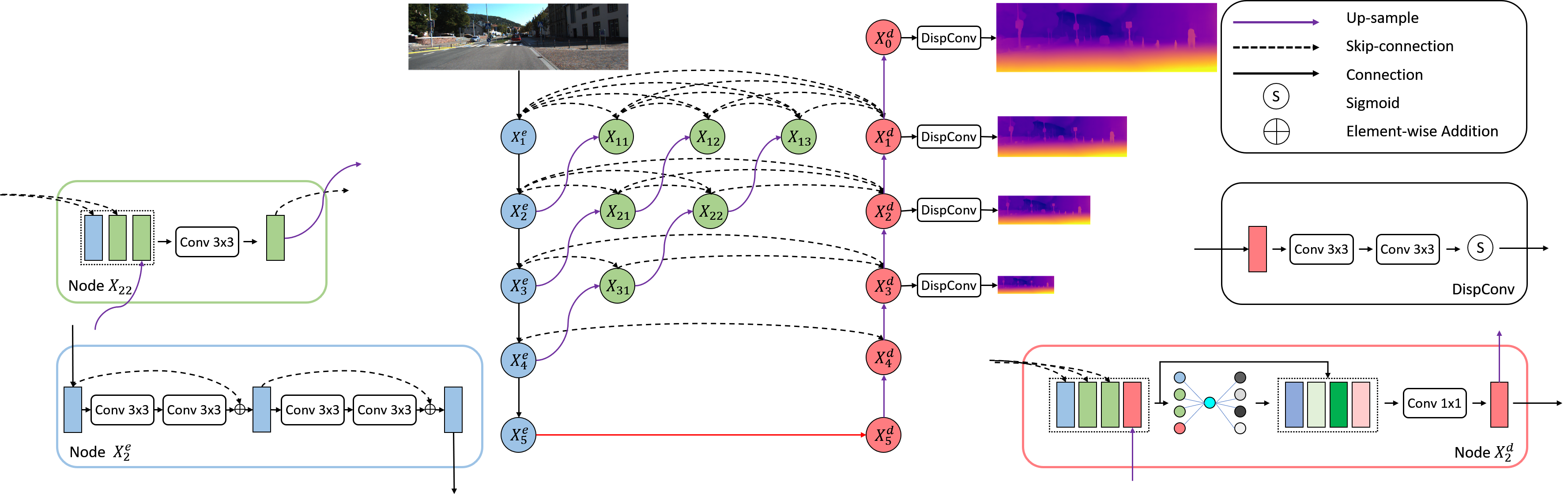} 
    \caption{\textbf{Illustration of our proposed framework.} The network is mainly composed of three different types of nodes. $X^{e}_i$ denotes a feature extraction node, which is mainly composed
    of residual blocks. $X_{i,j}$ denotes a feature fusion node which only has $3\times3$ convolution operation. $X^{d}_i$ is feature fusion node which is mainly composed of our proposed fSE module. Disparity is decoded
    by DispConv block, which contains $3\times3$ convolution and sigmoid activation function. }
\label{fig3}
\end{figure*}
\begin{table}[ht]
    \setlength{\tabcolsep}{3mm}{
    \begin{tabular}{lccc}
    \toprule
                        & Resolution          & Abs rel      &$\delta < 1.25$\\
    \hline
    Monodepth2          & $640\times192$      & 0.115        &  0.877         \\
    Monodepth2          & $1024\times320$     & 0.115        &  0.879         \\
    \bottomrule
    \end{tabular}
    }
    \caption{\textbf{Quantitative results from Monodepth2 with different resolution setting training and testing.}}
    \label{tab:1}
\end{table}

Furthermore, we make a deep analysis on the actual reason that existing methods cannot improve the depth estimation by high-resolution inputs. We find out that the most vital point is 
un-negligible error from the bilinear interpolation process for up-sampling the low resolution prediction to high ones. In detail, taking Monodepth2\cite{godard2019digging} as an example, 
when feeding a model with small inputs, we need to up-sample the outputs to obtain a required high-resolution one with interpolation. As shown in Figure 2, it can be seen that local 
prediction error will severely harm global accuracy in regions with large depth gradient such as instance edges, while for the region with small depth gradient. 
Interestingly, when the prediction of low-resolution outputs is poor like the upper right part in Figure 2, it will unexpectedly compensate the local prediction error of bilinear interpolation. 
The results show that the performance of low-resolution can be on par of the high one. In other words, the actual reason that most methods cannot benefit from larger inputs is that the 
gap between these two settings is remedied by the above interesting phenomenon. Therefore, only more precise prediction of regions with large depth gradient can make high-resolution prediction
more accurate. We can also summarize that the performance of high-resolution can be improved by predicting more accurate depth in large gradient regions and sharper edges.

\subsection{Redesign Skip Connection}
Based on the above analyses, in order to predict more accurate boundaries, we try to enhance it from both the spatial and semantic feature, since we believe (1) semantic information 
can produce boundaries between different categories, which can reduce depth estimation error caused by misclassification, (2) spatial information can help the network to know the location of
the boundaries, so as to estimate them better. Here, we will first discuss DepthNet with \textit{U-Net} architecture.

Skip-connection is one of core components of U-Net, whose purpose is to recover information lost in down-sampling. However, we argue that it could be less 
effective to directly combine features from different layers since there are gaps among them in semantic levels and spatial resolution. As previous research\cite{zhang2018exfuse} indicates, 
deep neural networks represent more semantic features as the layers going deeper. Thus, if low-level features could include more semantic information, like relatively clearer semantic boundaries, 
then the fusion becomes easier and we can obtain sharper depth estimation by decoding these features.

In order to decrease semantic and resolution gap, we propose dense skip connection inspired by \cite{zhou2018unet++}. As shown in Figure 3, in addition to the nodes in the original encoder and decoder,
we also add a lot of intermediate nodes to aggregate features. Let $x^{e}_i$ denotes the output of encoder node $X^{e}_i$, $x^{d}_i$ denotes the output of node $X^{d}_i$,
$x_{i,j}$ denotes the output of node $X_{i,j}$ where $i$ indexes the down-sampling layer along the encoder and j indexes the aggregation layer along the skip connection. Consider a single image $I$ is
passed through DepthNet. The stack of feature maps is computed as

\begin{align}
    &x^{e}_i = \left\{
    \begin{array}{ll}
        \mathcal{E}(I),&i = 1 \\
        \mathcal{E}(x^{e}_{i-1}),&i > 1 
    \end{array}
    \right.\\
    &x_{i,j}=\left\{
    \begin{array}{ll}
        \mathcal{F}(\Big[\big[x^{e}_{i}, [x_{i,k}]^{j-1}_0\big],\mathcal{U}(x_{i+1,j-1})\Big]),&j = 1 \\
        \mathcal{F}(\Big[\big[x^{e}_{i}, [x_{i,k}]^{j-1}_0\big],\mathcal{U}(x^{e}_{i+1})\Big]),&j > 1
    \end{array}
    \right. \\
    &x^{d}_{i}=\left\{
    \begin{array}{ll}
        \mathcal{D}(\mathcal{U}(x^d_{i+1})]),&j = 0  \\
        \mathcal{D}(\Big[\big[x^{e}_{i}, [x_{i,k}]^{j-1}_0\big],\mathcal{U}(x^d_{i+1})\Big]),&j > 0 
    \end{array}
    \right.
\end{align}

where $\mathcal{E}(\cdot)$ is a feature extraction block like residual block, $\mathcal{F}(\cdot)$ is a feature fusion block consist of convolution operation followed by an activation function, $\mathcal{D}(\cdot{})$ is a feature
fusion operation composed by feature fusion block, $\mathcal{U}(\cdot)$ is an upsampling block with convolution and bilinear interpolation operation, and $[\cdot]$ denotes
the concatenation layer. The details of dense skip-connection and each node are shown in Figure 3.

With dense skip connection, each node in the decoder is presented with the final aggregated feature maps, the intermediate aggregated feature maps and the original
feature maps from the encoder. Then, the decoder can use high-resolution features with richer semantic information to predict more sharper depth maps.

\subsection{Feature Fusion SE Block}
The U-Net based DepthNet uses a $3\times3$ convolution to fuse upsampling feature maps and original feature maps from the encoder, and the parameters of convolution is calculated as
\begin{equation}
    C_{in}\times C_{out}\times k^2  + C_{out},
\end{equation}
where $C_{in}$ is the number of input channels, $C_{out}$ is the number of output channels and $k$ denotes as convolution kernel size.
Just like \cite{huang2017densely}, the dense skip-connections make the decoder nodes have dramatically increased input feature maps, so this operation will inevitably decrease the efficiency of the network. 
Inspired by \cite{hu2019squeeze-and-excitation}, we propose a lightweight module, feature fusion Squeeze-Excitation(fSE), to improve feature fusion accuracy and efficiency. The fSE module squeezes
the feature maps by global average pooling to represent channel information and uses two fully-connected(FC) layers followed by a sigmoid function to measure the importance of each feature and re-weight
them in the meantime. Then a $1\times1$ convolution is used to fuse channels to obtain high-quality feature maps. The parameter amount of this module is
\begin{equation}
    \frac{2}{r}\times C_{in}^2 + (C_{in} + 1)\times C_{out},
\end{equation}
where $r$ denotes as reduction ratio and is always set to 4 in all experiments of this paper. When using Resnet-18 as encoder, the fSE module can reduce the parameters of HR-Depth from 16.06M to 14.62M, even less 
than Monodepth2 with 14.84M parameters. Meanwhile, since the fSE module will focus more on feature fusion, the network performance will also be improved.

\begin{table*}
    \centering
    \setlength{\tabcolsep}{1mm}{
    \begin{tabular}{lcccccccccc}
    \toprule
                                    &Supervision& Resolution      &Dataset  & Abs rel & Sq Rel & RMSE   & RMSE$_{log}$ &$\delta < 1.25$ &$\delta < 1.25^2$ &$\delta < 1.25^3$\\
    \hline
    SfMLearner                      & M         & $416\times128$  & CS + K  & 0.198   & 1.836  & 6.565  & 0.275    &  0.718         & 0.901            & 0.960\\
    Vid2Depth                       & M         & $416\times128$  & CS + K  & 0.159   & 1.231  & 5.912  & 0.243    &  0.784         & 0.923            & 0.970\\
    Struct2Depth                    & M         & $416\times128$  & K       & 0.141   & 1.026  & 5.291  & 0.215    &  0.816         & 0.945            & 0.979\\
    Monodepth2                      & M         & $640\times192$  & K       & 0.115   & 0.903  & 4.863  & 0.193    &  0.877         & 0.959            & 0.981\\
    PackNet-SfM                     & M         & $640\times192$  & K       & 0.111   & \textbf{0.785}  & \textbf{4.601}  & 0.189    &  0.878         & 0.960            & 0.982\\
    HR-Depth\textbf{(Ours)}        & M         & $640\times192$  & K       & 0.109   & 0.792  & 4.632& \textbf{0.185}    &  0.884         & \textbf{0.962}            & \textbf{0.983}\\
    HR-Depth\textbf{(Ours)}        & M         & $640\times192$  & CS + K  & \textbf{0.108}   & 0.955  & 4.800& 0.190    &  \textbf{0.887}         & 0.961            & 0.981\\
    \hline
    Monodepth2                      & MS        & $640\times192$  & K       & 0.106   & 0.818  & 4.750  & 0.196    &  0.874         & 0.957            & 0.979\\
    HR-Depth\textbf{(Ours)}        & MS        & $640\times192$  & K       & 0.107   & 0.785  & 4.612  & 0.185    &  0.887         & 0.962            & 0.982\\
    HR-Depth\textbf{(Ours)}        & MS        & $640\times192$  & CS + K  & \textbf{0.104}   & \textbf{0.786}  & \textbf{4.544}  & \textbf{0.182}    &  \textbf{0.893}         & \textbf{0.964}            & \textbf{0.983}\\
    \hline
    Zhou et al.                     & M         & $1248\times384$ & K       & 0.121   & 0.837  & 4.945& 0.197    &  0.853         & 0.955            & 0.982\\
    Monodepth2                      & M         & $1024\times320$ & K       & 0.115   & 0.882  & 4.701& 0.190    &  0.879         & 0.961            & 0.982\\
    PackNet-SfM                     & M         & $1280\times384$ & K       & 0.107   & 0.802  & 4.538& 0.186    &  0.889         & 0.962            & 0.981\\
    HR-Depth\textbf{(Ours)}        & M         & $1024\times320$ & K       & 0.106   & 0.755  & 4.472& 0.181    &  0.892         & \textbf{0.966}            & \textbf{0.984}\\
    HR-Depth\textbf{(Ours)}        & M         & $1280\times384$ & K       & \textbf{0.104}   & \textbf{0.727}  & \textbf{4.410}& \textbf{0.179}    &  \textbf{0.894}         & \textbf{0.966}            & \textbf{0.984}\\
    \hline
    Monodepth2                      & MS        & $1024\times320$ & K       & 0.106   & 0.818  & 4.750& 0.196    &  0.874         & 0.957            & 0.979\\
    HR-Depth\textbf{(Ours)}        & MS        & $1024\times320$ & K       & \textbf{0.101}   & \textbf{0.716}  & \textbf{4.395}& \textbf{0.179}    &  \textbf{0.899}         & \textbf{0.966}            & \textbf{0.983}\\

    \bottomrule
    \end{tabular}}
    \caption{\textbf{Quantitative results of depth estimation on KITTI dataset for distance up to 80m.} For error evaluating indexes, Abs Rel, Sq Rel, RMSE and RMSE$_log$, lower is better, and 
    for accuracy evaluating indexes, $\delta < 1.25, \delta < 1.25^2, \delta < 1.25^3$, higher is better. In the dataset column, CS + K refers to pre-training on CityScapes(CS) and fine-tuning on KITTI(K).
    M refers to DepthNet that is supervised by monocular(M) image sequence and MS refers to DepthNet that is supervised by monocular and stereo (MS) images. At test time, we scale outputs of DepthNet with
    median ground-truth LiDAR information.}
    \label{tab:2}
\end{table*}

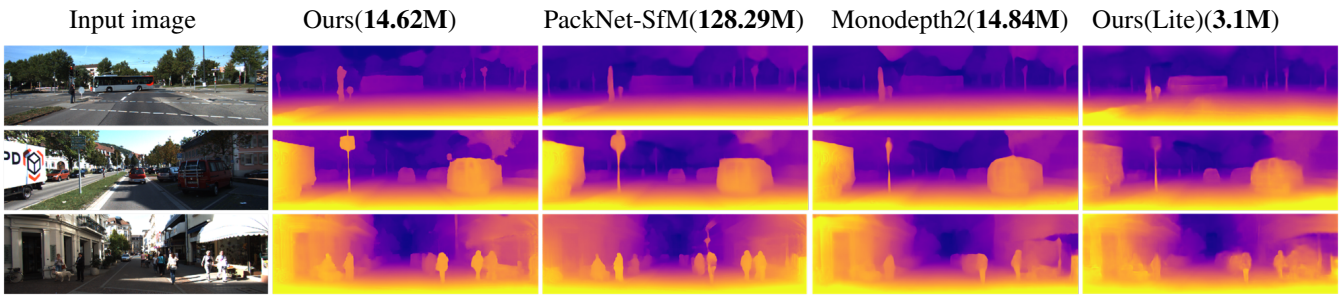
\begin{figure*}[!ht]
    \centering
    \begin{overpic}[width=1\textwidth]{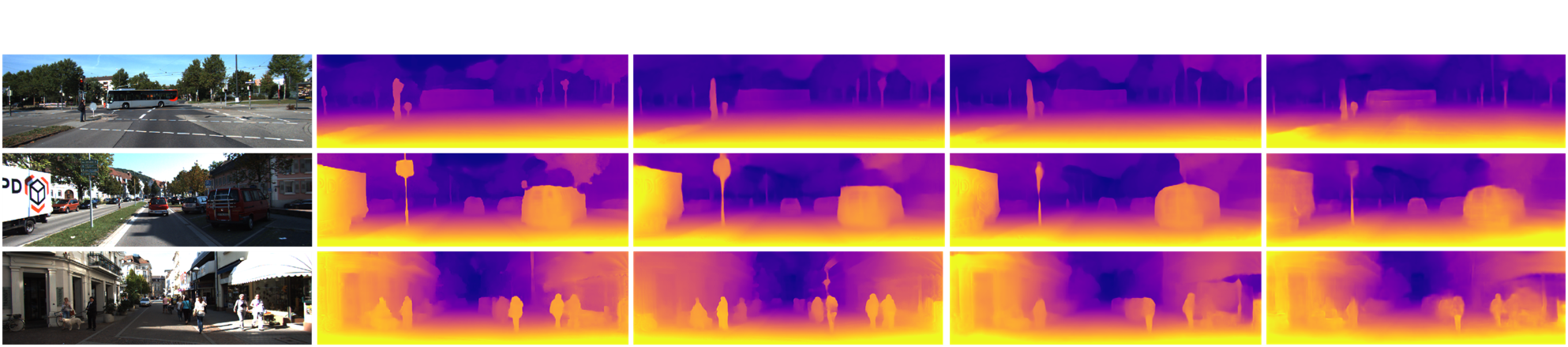}
        \put(5,20){Input image}
        \put(22.5,20){Ours(\textbf{14.62M})}
        \put(40.5,20){PackNet-SfM(\textbf{128.29M})}
        \put(62,20){Monodepth2(\textbf{14.84M})}
        \put(81.5,20){Ours(Lite)(\textbf{3.1M})}
    \end{overpic}
    \caption{\textbf{Qualitative monocular depth estimation performance} comparing HR-Depth and Lite-HR-Depth with previous SOTA. on frames from the KITTI dataset. Our network is able to predict more sharper 
    edges than Monodepth2\cite{godard2019digging}, and its performance is comparable to PackNet-SfM while with much fewer parameters.}
\label{fig4}
\end{figure*}

\subsection{Lite-HR-Depth}
Most previous SoTA self-supervised monocular depth estimation algorithms are based on fairly complex deep neural networks with a mass of parameters which limit their
applications on practical platforms like embedded devices. However, for existing lightweight networks, such as PyD-Net\cite{poggi2018towards}, the accuracy is greatly decreased when they reduce the 
amount of parameters. In order to make the depth estimation network get rid of the limitation of GPU but keep its great performance, we propose a simple yet effective lightweight network, named Lite-HR-Depth. 

Our Lite-HR-Depth employs MobileNetV3 as an encoder and downsizes the feature fusion and decoder nodes with only 3.1M parameters, of which the 2.82M parameters comes from the encoder. In addition, we further
improve the accuracy of our Lite-HR-Depth with knowledge distillation\cite{hinton2015distilling}. For self-supervised learning, due to lack of ground truth, we have to use view syntheses as the supervisory signal, which increases
the difficultly of training small network. Therefore, we propose to source a direct form of supervision from the learned large model.
By training a large network in a self-supervised manner, we obtain a high-performance network instance $\mathcal{T}$. Then we train a second instance of lightweight model, namely $\mathcal{S}$, to minimize 
\begin{equation}
    L_{sup} = ||d_{\mathcal{T}} - d_{\mathcal{S}}||,
\end{equation}
where $d_{\mathcal{T}}$ means disparity from network $\mathcal{T}$ and $d_{\mathcal{S}}$ means disparity from network $\mathcal{S}$. With this strategy, we obtain a network $\mathcal{S}$ 
which even more accurate than $\mathcal{T}$. It is worth noting that the parameters of Lite-HR-Depth is only $20\%$ of Monodepth2, but it can even perform better than Monodepth2 
at high resolution. More analyses of this lite network will be shown in our experiment.

\section{Experiments}
In this section, we validate that (1) our redesigned skip-connection can improve the results, especially predicting sharper edges, (2) the fSE module can significantly reduce parameters and improve accuracy, and
(3) the design method we propose can easily obtain high precision lightweight network. We evaluate our models on the KITTI dataset\cite{geiger2013vision}, to allow comparison with previous published monocular methods.
\subsection{Datasets}
\textbf{KITTI.} The KITTI benchmark \cite{geiger2013vision} is most widely used for depth evaluation. 
We adopt the data split of \cite{eigen2015predicting}, and removed the static frames followed by \cite{zhou2017unsupervised}. Ultimately, we used 39810 images for training, 4424 for validation and 697 for evaluation. 
Furthermore, we use the same intrinsic for all images, setting the principal point of the camera to the image center and the focal length to the average of all the focal lengths in KITTI. For stereo training, we set the transformation
between the two stereo frames to be a pure horizontal translation of fixed length.

\noindent\textbf{CityScapes.} CityScape \cite{Cordts2016Cityscapes} is the other large automatic driving dataset. So we also experiment with pre-training our structure on CityScape and then we finetuned and evaluated it on KITTI dataset. 
The \textbf{leftImg8bit Sequence} were considered as training split for the CityScapes dataset, using the same training parameters as KITTI for 20 epochs.

\subsection{Implementation Details}
We implement our models on PyTorch\cite{paszke2017automatic} and train them on one Telsa V100 GPU. We use the Adam Optimizer\cite{kingma2014adam} with $\beta_1 = 0.9, \beta_2 = 0.999$. The DepthNet and PoseNet are 
trained for 20 epochs, with a batch size of 12. The initial learning rates for both network are $1\times10^{-3}$ and decayed after 15 epochs by factor of 10. The training sequences are consist of three consecutive 
images. We set the SSIM weight to $\alpha=0.85$ and smooth loss weight to $\lambda=1\times10^{-3}$.

\noindent\textbf{DepthNet.} We implement our HR-Depth with ResNet-18\cite{Resnet} as encoder, and use MobileNetV3\cite{howard2019searching} as encoder for Lite-HR-Depth. The details of our proposed architecture will be described at supplemental material. 
All four disparity maps are used in loss calculation during training. For evaluation, we only use the maximum output scale, after being resized to the ground-truth depth resolution using bilinear interpolation. 

\noindent\textbf{PoseNet.} The architecture of PoseNet is proposed by \cite{godard2019digging}. The PoseNet is built on Resnet-18 and the first-level convolution channel is changed from 3 to 6, which allows the adjacent frames to feed into the network. 
And the outputs of PoseNet is the relative pose which is parameterized with 6-DOF vector. The first three dimensions represent translation vectors and the last three represent Euler angles.

\subsection{Depth Estimation Performance}
We evaluate depth prediction on KITTI using the metrics described in \cite{eigen2015predicting}. The evaluation results are summarized in Table 2 and then illustrate their performance qualitatively in Figure 4. 
We show that our proposed architecture outperforms all existing SoTA self-supervised approaches. We also outperform recent model \cite{packnet} with $120M$ parameters. Furthermore, we also introduce 
an additional source of unlabeled videos, CityScapes dataset(CS+K), and we can further improve the DepthNet performance. We also show that at higher resolution out model's performance significantly increases.
Our best results are achieved SoTA when processing higher resolution input images with minimum parameters. 

\noindent As shown in Table 3, when training with low-resolution images, Lite-HR-Depth can perform better than Monodepth2 with teacher (T) network supervising. However, when training with high-resolution images, 
Lite-HR-Depth can perform better than Monodepth2 without additional supervision signal (M).

\begin{table*}[!ht]
    \centering
    \setlength{\tabcolsep}{0.7mm}{
    \begin{tabular}{lcccccccccc}
    \toprule
                                    &Supervision& Resolution      &$\#Para$  & Abs rel & Sq Rel & RMSE   & RMSE$_{log}$ &$\delta < 1.25$ &$\delta < 1.25^2$ &$\delta < 1.25^3$\\
    \hline
    PyD-Net                         & M         & $512\times256$  & 1.9M    & 0.146   & 1.291  & 5.907  & 0.245    &  0.801         & 0.926            & 0.967\\
    Monodepth2                      & M         & $640\times192$  & 14.84M  & 0.115   & 0.903  & 4.863  & 0.193    &  \textbf{0.877}         & 0.959            & 0.981\\
    Lite-HR-Depth(\textbf{Ours})   & M         & $640\times192$  & 3.1M    & 0.116   & 0.845  & 4.841  & 0.190    &  0.866         & 0.957            & 0.982\\
    Lite-HR-Depth(\textbf{Ours})    & T         & $640\times192$  & 3.1M    & \textbf{0.112}   & \textbf{0.767}  & \textbf{4.557}  & \textbf{0.184}    & 0.876         & \textbf{0.963}            & \textbf{0.984}\\
    \hline
    Monodepth2                      & M         & $1024\times320$ & 14.84M  & 0.115   & 0.882  & 4.701  & 0.190    &  0.879         & 0.961            & 0.982\\
    Lite-HR-Depth(\textbf{Ours})   & M         & $1024\times320$ & 3.1M    & 0.111   & 0.799  & 4.612  & 0.184    &  0.878         & 0.963   & 0.984\\
    Lite-HR-Depth(\textbf{Ours})  & T         & $1024\times320$ & 3.1M    & \textbf{0.105}   & \textbf{0.704}  & \textbf{4.374}  & \textbf{0.178}    &  \textbf{0.891}   & \textbf{0.967}   & \textbf{0.985}\\
    \hline
    Lite-HR-Depth(\textbf{Ours})   & T         & $1280\times384$ & 3.1M    & \textbf{0.104}   & \textbf{0.697}  & \textbf{4.349}  & \textbf{0.176}    &  \textbf{0.893}   & \textbf{0.967}   & \textbf{0.985}\\
    \bottomrule
    \end{tabular}}
    \caption{\textbf{Quantitative performance of Lite-Network on KITTI dataset for distance up to 80m.} Evaluation matrices and methods are the same as Table 2. In supervision column, T refers to using teacher network to guide
    lite network to train. }
    \label{tab:3}
\end{table*}

\begin{figure}[!ht]
    \centering
    \includegraphics[width=1\columnwidth]{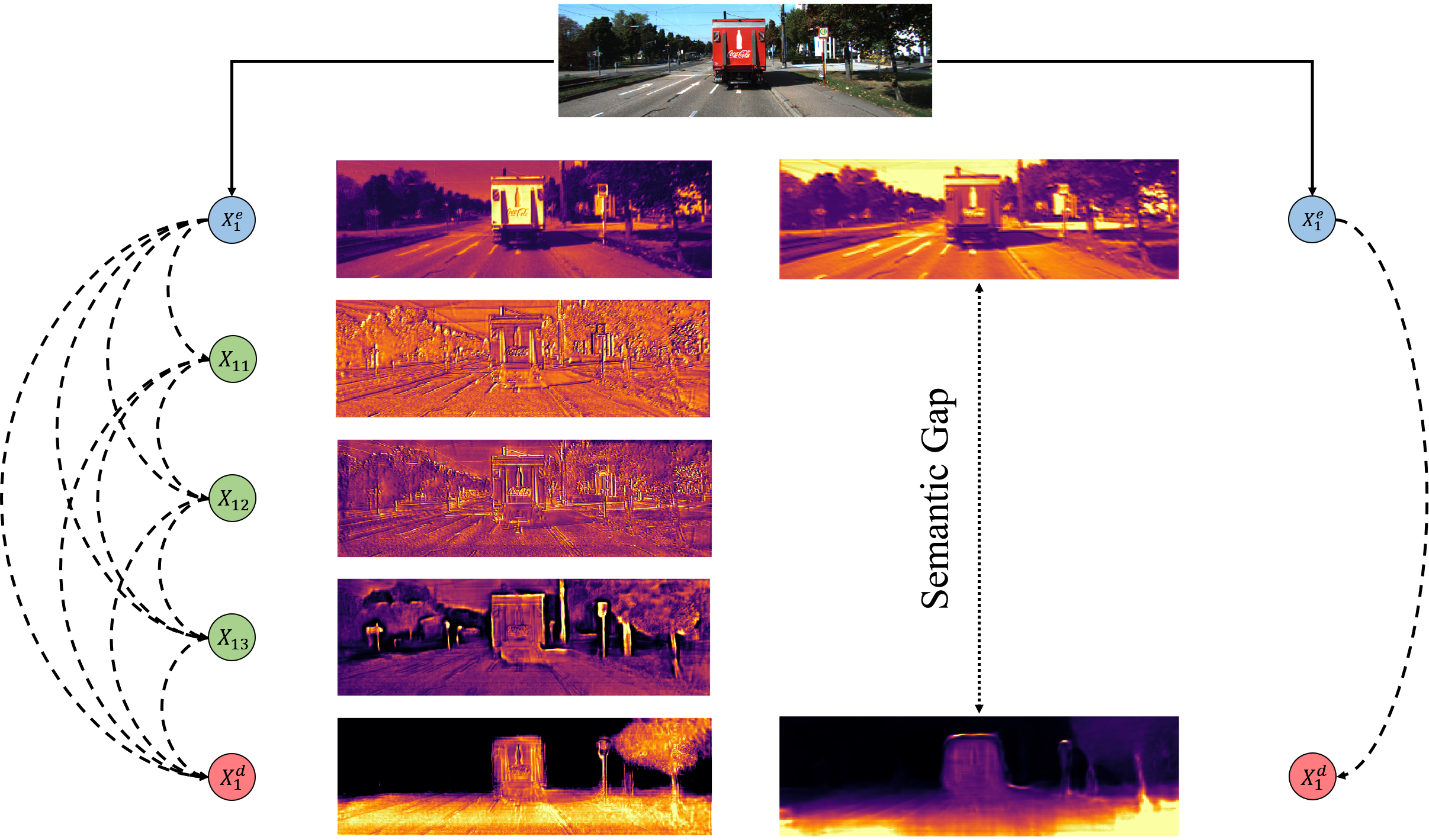} 
    \caption{\textbf{Visualization of feature maps at first level} Left side shows feature maps in HR-Depth and right side shows feature maps in Monodepth2.}
\label{fig5}
\end{figure}

\subsection{Ablation Studies}
To further explore the performance improvements that our network provides, we perform an ablative analysis on the different architectural components introduced. We choose Monodepth2 as our baseline model, and we can see 
all our contributions can lead a significant improvement. 

\noindent\textbf{Benefits of dense skip-connection.} As shown in the Figure 5, the semantic gap in skip connection is too large to be fused well, but the dense skip-connection can leverage intermediate features to 
effectively reduces the semantic gap between encoder and decoder. Therefore, we can get high-resolution feature maps with more semantic information which can significantly improve the performance. 

\noindent\textbf{fSE Block.} The fSE block is designed to improve feature fusion and computation efficiency. To validate fSE, we also apply the SE\cite{hu2019squeeze-and-excitation} to dense skip connection and 
compare it with the proposed fSE block. As shown in Table 4, the SE block will introduce additional parameters but our fSE block can greatly reduce the
parameters introduced by dense skip-connection and can further improve the performance of the network, even better than SE block.

\begin{table}[!ht]
    \setlength{\tabcolsep}{1mm}{
    \begin{tabular}{lcccc}
    \toprule
                        & $\#Para$  & Abs rel   & RMSE &$\delta < 1.25$\\
    \hline
    baseline            & 14.84M     & 0.115  & 4.863        &  0.877         \\
    +dense SC           & 16.06M     & 0.112  & 4.706        &  0.881         \\
    +fSE                & 13.48M     & 0.114  & 4.821        &  0.875         \\
    +dense SC and SE    & 16.23M     & 0.110  & 4.655        &  0.882         \\
    +dense SC and fSE   & 14.62M     & \textbf{0.109}  & \textbf{4.632}        &  \textbf{0.884}         \\
    \bottomrule
    \end{tabular}
    }
    \caption{\textbf{Ablation Studies.} Results for different variants of our model with monocular training on KITTI at low resolution on Eigen split. The baseline model is Monodepth2. We introduce the dense skip-connection(dense SC)
    to original structure and compare effect of SE block and fSE block.}
    \label{tab:4}
\end{table}

\subsection{Feature map visualization}
As explained before, the purpose of redesigning skip-connection is to decrease semantic and spatial gap, so as to obtain high-resolution feature maps with rich semantic information. Therefore, in order to illustrate
the effects of feature fusion, we visualize the intermediate feature maps and we also plot the outputs of fSE to explore the influence of intermediate feature maps. Figure 4 shows that the low-level node like $X^{e}_1$ undergoes slight
transformation and obtains simple spatial information whereas the output of decoder node like $X^{d}_2$ gets the rich semantic information. Hence, there is large gap between the representation capability of $X^{e}_1$ and $X^{d}_2$.
The dense skip connection can gradually add semantic information to the intermediate feature, thus reducing the gap between node $X^{e}_1$ and $X^{d}_2$. 

\section{Conclusion}
In this paper, we show theoretical and empirical evidence that how to improve high-resolution estimation performance. And based on analysis, we present a new convolutional network architecture, referred as HR-Depth, for high-resolution 
self-supervised monocular depth estimation. It leverages novel dense skip-connection and fSE block to reduce the gap between resolution and semantic. Although purely trained on image sequences, out approach outperforms other existing 
self and semi-supervised methods and is even competitive with supervised method. Furthermore, we propose a simple yet efficient strategy to design the lightweight network. The experiments demonstrate that Lite-HR-Depth can perform on
par with large model with fewer parameters.

\section{Acknowledgments}
This work is supported in part by the Key Research and Development Program of Guangdong Province of China (2019B010120001) and the National Natural Science Foundation of China under Grant 61836015.
\bibliography{
    aaai21
}
\end{document}